\documentclass{ceurart}

\usepackage{todonotes}

\usepackage{adjustbox}

\usepackage{amsmath}

\usepackage{tabularx}
\newcolumntype{Y}{>{\centering\arraybackslash}X}

\usepackage[ruled,vlined]{algorithm2e}

\usepackage{subcaption}

\usepackage{booktabs}

\usepackage{array}
\newcolumntype{L}{>{\centering\arraybackslash}m{2cm}}

\newcommand{\partitle}[1]{\vspace{1mm}\noindent\textbf{#1}}

\usepackage{url}
\usepackage{graphicx}

\begin{document}

\copyrightyear{2021}
\copyrightclause{Copyright for this paper by its authors.
  Use permitted under Creative Commons License Attribution 4.0
  International (CC BY 4.0).}

\conference{IIR'21: Italian Information Retrieval Workshop (IIR)}

\title{Keyword Extraction for Improved Document Retrieval in Conversational Search}

\author[1]{Oleg Borisov}[%
email=oleg.borisov@usi.ch,
]
\address[1]{Universit\'a della Svizzera italiana (USI), Lugano, Switzerland}

\author[2]{Mohammad Aliannejadi}[%
email=m.aliannejadi@uva.nl,
url=http://aliannejadi.com
]
\address[2]{University of Amsterdam, Amsterdam, Netherlands}

\author[1]{Fabio Crestani}[%
email=fabio.crestani@usi.ch,
]

\begin{abstract}
Recent research has shown that mixed-initiative conversational search, based on the interaction between users and computers to clarify and improve a query, provides enormous advantages. Nonetheless,  incorporating additional information provided by the user from the conversation poses some challenges. In fact, further interactions could confuse the system as a user might use words irrelevant to the information need but crucial for correct sentence construction in the context of multi-turn conversations.
To this aim, in this paper, we have collected two conversational keyword extraction datasets and propose an end-to-end document retrieval pipeline incorporating them. Furthermore, we study the performance of two neural keyword extraction models, namely, BERT and sequence to sequence, in terms of extraction accuracy and human annotation. Finally, we study the effect of keyword extraction on the end-to-end neural IR performance and show that our approach beats state-of-the-art IR models. We make the two datasets publicly available to foster research in this area.
\end{abstract}


\maketitle              %

\section{Introduction}

Recent developments in speech recognition and deep learning have led to intelligent assistants, such as Google Assistant, Microsoft Cortana, and Apple Siri. Consequently, researchers and users are exploring novel means of communication and information access, such as spoken queries and conversations~\cite{DBLP:conf/chiir/RadlinskiC17}. Research on information-seeking conversational systems has gained lots of attention recently. Various shared evaluation tasks have been raised in the community, focusing on single-~\cite{DBLP:conf/sigir/0001XKC20} and mixed-initiative~\cite{DBLP:journals/corr/abs-2009-11352,alian21,aliannejadi21buidling} conversational search systems. The aim of research in mixed-initiative conversations is to enable a system to take the initiative of the conversation when necessary, aiming to provide a better experience to the user~\cite{DBLP:conf/chi/Horvitz99}. An example of mixed-initiative interaction is asking clarifying questions that has been recently studied in the context of information-seeking conversations~\cite{DBLP:conf/sigir/AliannejadiZCC19,DBLP:conf/sigir/HashemiZC20} and Web search~\cite{DBLP:conf/www/ZamaniDCBL20,DBLP:conf/sigir/ZamaniMCLDBCD20,DBLP:conf/ecir/SekulicAC21,DBLP:lotze21}.

In Web search, where users usually type their queries, they take some time to formulate a query and often do not follow common sentence structures. For example, they only focus on using the most important words for their search. Consequently, a narrow focus is created for the search engine, making the inspection of documents for the most relevant query words easier.
In contrast to this, conversational IR faces challenges due to the inclination of users to follow their own speech patterns when formulating queries rhetorically. Here, users tend to include some unnecessary terms that appear crucial for a proper sentence construction but might derail the IR model in searching for relevant documents~\cite{kato2014cognitive}. This could also be magnified when a conversation evolves into multiple turns~\cite{DBLP:conf/chiir/AliannejadiCRC20,DBLP:conf/sigir/VoskaridesLRKR20} and a new form of conversation is presented to the user, such as when the system asks clarifying questions. This happens mainly due to the context-dependence nature of multi-turn conversations and new types of responses that could emerge in a mixed-initiative conversation.

While the effectiveness of conversational systems has been studied before \cite{DBLP:conf/sigir/AliannejadiZCC19,zhang2018towards,DBLP:conf/sigir/HashemiZC20,DBLP:conf/ictir/KrasakisAVK20}, the main goal of this paper is to study if the identification of keywords retrieved from the human-computer interaction will help achieve better retrieval results. To this aim, we collect two datasets of keyword extraction and study the effectiveness of multiple generative models on them. Our first dataset is collected based on the performance of the retrieval model using different keywords, while the other  is collected from news articles online. Every news article comes with a title and a set of keywords. Our intuition is that a neural model can learn to extract useful keywords from news titles and use this external knowledge for more effective keyword extraction in a conversation. We study the effect of various keyword extraction strategies on non-neural and neural document retrieval pipelines.
To the best of our knowledge, keyword extraction in the context of mixed-initiative conversational IR has not been studied before.

In our retrieval pipeline, after the conversational phase, where the system interacts with the user to clarify the query ambiguities, the conversational sequence is passed to the keyword extractor, identifying the most important terms from the sentences. In parallel to that, the document retrieval model performs the first relevance ranking of the documents based on the \textbf{original} conversation. Finally, the Neural IR performs re-ranking using top documents from the \textit{IR phase 1} and \textbf{keywords} obtained by keyword extractor from \textit{Keyword Extraction Phase} as inputs of the system.

\section{Data Collection} \label{sec:data}

As the topics addressed by this study have only recently surfaced in research, a substantial amount of work needed to be done to answer whether keywords could support the IR model with document retrieval tasks. 

As was discussed earlier, keyword extraction from short-sized documents using Deep Learning is a relatively new topic. The previously created Inspec, SemEval-2010, SemEval-2017 datasets are not suitable for this research, as they are focused on keyword and keyphrase extraction from medium- and large-sized texts (e.g., abstracts or scientific articles)~\cite{augenstein2017semeval,hulth2003improved,kim2013automatic}.
In contrast to this, the main focus of this research is keyword extraction from small-sized sentences of the length of no more than 20 words, which is the average English sentence length~\cite{cutts2020oxford}.
Therefore, we collect and release two types of datasets: (i) News-Keyword based and (ii) IR-Keyword based datasets.\footnote{Data available at: \url{https://github.com/aliannejadi/ConvKey}}

\subsection{News-Keywords Based dataset}
Online newspaper websites and other social network Web pages tend to follow a content structure, where common articles are structured as title, main text, tags (sometimes hashtags). Content creators try not only to select an appealing and interesting title name but also to summarize the content in one sentence, thus selecting the most important words to portray the key message of an article. This can also be considered a reverse  IR operation as the author, given the document's content, provides a title (considered the \textit{query} in our case) that corresponds to the article in the best possible way. 

Authors usually also choose some tags that either describe the article in the most general way or place the story in the context of other related articles that one could find on the website. From the user’s point of view, tags provide an opportunity to navigate to other related material; however, having well-formulated tags is also crucial for Search Engine Optimization and could impact the website's visibility or the article \cite{yalccin2010search}.

Taking into consideration that writers pay very close attention to the title and the tags used,
where it is not unusual for tag words to appear in the title, brings us to the first method of
keyword dataset: considering \textit{title} as the \textbf{input text}, and \textit{tags} as the \textbf{target keywords}. If a tag does not appear in a corresponding title, we do not add it to the keywords list (as shown in Table \ref{tab:news_dataset_example}).

\begin{table}[!ht]
\vspace{-3mm}
    \caption{Example of scraped news article\protect\footnotemark}
    \centering
    \begin{tabular}{ll@{\quad}l}
    \toprule
     \multicolumn{1}{c}{\textbf{Title}} & \multicolumn{1}{c}{\textbf{Tags}}  & \multicolumn{1}{c}{\textbf{Keywords List}}  \\
     \midrule
        Five German words you'll need \hspace{1mm}  & [\textit{summer}, holidays, members] &  [\textit{summer}] \\ to know this \textit{summer} &  & \\
    \bottomrule
    \end{tabular}
    \vspace{-5mm}
    \label{tab:news_dataset_example}
    
\end{table}

\footnotetext{Taken on 30 June 2020 from \\ \url{https://www.thelocal.ch/20200630/five-german-words-youll-need-to-know-this-summer}}

To create the dataset, we scraped the following news websites: BBC\footnote{https://www.bbc.com/}, The Local\footnote{https://www.thelocal.ch/} and Salon\footnote{https://www.salon.com/}. In total, over 104,000 title-tag pairs have been obtained using this method. After filtering the outliers and the items where the tags do not appear in the title, the dataset shrinks to 79,000 instances.

\subsection{IR-Keyword-based dataset}
Classical IR systems only focus on the basic preprocessing of the query, such as the removal of stopwords and punctuation. Having too many words could confuse the system and lead it to retrieve unwanted results. Therefore, a correct keyword identification could lead to better retrieval performance, while selecting less good keywords will inevitably worsen the output results.
We developed the IR-Keyword-based dataset based on this assumption, applying the previously created Qulac dataset~\cite{DBLP:conf/sigir/AliannejadiZCC19}.
To create a dataset in this context, we used Qulac's first conversational round, containing three components: query $q$, question $t$, and answer $a$, which retrieve a set of relevant documents.

The main idea is to identify a set of words from $q, t$, and $a$, which will lead to the greatest relevance of retrieved documents. To evaluate the system's performance, we used the Normalized Discounted Cumulative Gain at 20 (NDCG@20) metrics. 
Due to the complexity of the permutations of all potential keywords of the whole set $q, t, a$, we decided to focus on one component at a time. The algorithm that was used is presented in Algorithm \ref{alg:kwds_pseudocode}. The main idea is to choose $s_0$, which could be a query, question, or an answer. For example, let us consider $s_0$ a \textit{query} and $s_1, s_2$ to represent the \textit{question} and \textit{answer}. 
Afterward, we would like to consider all possible subsets of words of $s_0$ (query in our example), which will form a set of \textit{potential keywords}. In mathematical terms, such operation is known as \textbf{the powerset}. For instance, if $s_0$ is "How are you?", then a set of \textit{potential keywords} would be: \{"how", "are", "you", "how are", "how you", "you are", "how are you" \}\footnote{We also keep the original order in which the words appear in the text}. The cardinality of a powerset highly depends on the number of words that the input sentence contains. To address having a large powerset, we limit the maximum size of the subset to four words. 

Next, we consider one instance $k_i$ from the \textit{potential keyword} set and retrieve the documents by supplying $k_i, s_1$ and $s_2$ to the document retrieval model. Consequently, it is important to evaluate the retrieved documents' relevance and save the obtained score. In the end, we save the $s_0$ as the \textbf{input text} and $k_{best}$ as the set of \textbf{keywords} that led to the retrieval of the most relevant documents.
We repeat a similar operation by considering $s_0$ as the question, and $s_1, s_2$ as query and question, and later $s_0$ as the answer, and $s_1, s_2$ as query and answer, respectively.
We apply the same process for all conversations from the Qulac dataset until we receive keywords from all queries, questions, and answers. Applying this approach, 15,320 data samples were obtained.
The benefit which this approach suggests is that where the answer of a user in a computer interaction is uncertain or ambiguous and will not provide any important information, the system learns to ignore these. In this scenario, the system should ideally ask another question or base the search only on the initial query. Therefore, the proposed method of dataset generation will be able to mimic this behavior.

\begin{algorithm}[t!]
\SetAlgoLined
\textbf{Method:} find\_keywords(query, question, answer)\\
 \For{$s_0$ \textbf{in} [query, question, answer]}{
    $s_1, s_2$ = [query, question, answer].remove($s_0$)\;
    potential\_keywords = PowerSet($s_0$, maxSubsetSize=4)  \;
    \vspace{2.5mm}

    scores\_list = list() \; %
    \For{$k_i$ \textbf{in} potential\_keywords}{
        ranked\_documents = IRmodel.retrieve($k_i, s_1, s_2$) \;
        score = ranked\_documents.evaluate(metrics="NDCG@20")\; 
        scores\_list.append(score) \;
    }
    max\_score\_index = \textbf{argmax}(scores\_list) \;
    $k_{best}$ = potential\_keywords[max\_score\_index] \; %
    save("input text" = $s_0$, "keywords"=$k_{best}$)

 }
 \caption{IR-Keyword Based Dataset Creation Method.}
 \label{alg:kwds_pseudocode}
\end{algorithm}

\section{Proposed Methods}
This section describes our conversational IR framework. We start with the neural models that we used for the keyword extraction task. Then we continue with the neural IR models and describe how keyword extraction fits into our pipeline.

\subsection{Keyword Extraction Models}

For the \textit{Keyword Extraction Phase}, we experimented with two different types of neural models: \textit{Sequence-to-Sequence} architecture and \textit{BERT} model~\cite{sutskever2014sequence,devlin2018bert}. 
Sequence-to-Sequence architecture uses Gated Recurrent Unit (GRU) as a recurrent neural network, the Attention mechanism to help the decoder, and pre-trained Word2Vec embeddings the performance on the words outside of the training set vocabulary \cite{bahdanau2014neural,cho2014learning,mikolov2013efficient}. We use Sequence-to-Sequence because it has been a state-of-the-art architecture for many different NLP tasks and established new benchmarks for the tasks of Neural Machine Translation \cite{sutskever2014sequence}.
In contrast to the previously described model, we also selected BERT as the most recently developed Transformer-based neural architecture in the field of NLP. One of its biggest advantages is that it has been pre-trained on a great amount of data using two main approaches: Masked Language Model, which is related to the prediction of masked/hidden tokens in the input sentence, and Next Sentence Prediction, which has the objective of predicting the next sentence from the input sequence. Therefore, by fine-tuning the model, it is possible to achieve great results in tasks, such as: Named Entity Recognition (NER), Sentence Classification, Answer Searching, and others \cite{devlin2018bert}.

To train selected architectures, we formulate the task of keyword extraction in the form of a NER task, as shown in Table \ref{tab:data_converted}. Where we say that a word is a \textbf{keyword} it its corresponding entity is labeled as "\textbf{1}", and \textbf{not a keyword} if it is marked as "\textbf{0}".

\begin{table}[t]
    \centering
    \vspace{-3mm}
    \caption[Example of Data Supplied to Neural Networks]{Example of Data Supplied to Neural Networks.\protect\footnotemark}
     \begin{tabular}{l@{\quad}l} 
        \toprule    
         \textbf{Original Sentence} &  ``Conservatives and liberals drink different beer'' \\ [0.5ex] 
         \textbf{Tokenized Sentence} & ['\textbf{conservatives}', 'and', '\textbf{liberals}' \\& \'drink', 'different', '\textbf{beer}']\\ 
         
         \textbf{Keywords} & ['conservatives', 'liberals', 'beer'] \\
         
         \textbf{Named Entities} & [\textbf{1}, 0, \textbf{1}, 0, 0, \textbf{1}] \\
         \bottomrule
         
    \end{tabular}
    \vspace{-5mm}
    \label{tab:data_converted}
    
    \end{table}

\footnotetext{ \small Retrieved on 15th of October 2019, from: \\ \url{https://www.salon.com/2013/02/27/conservatives_and_lilberals_drink_different_beer_partner/}}

\subsection{Neural IR models}
We extend the solution available from previous research~\cite{DBLP:conf/sigir/AliannejadiZCC19} by adding \textit{Information Retrieval Phase 2}, represented by the Neural IR model. We study the effectiveness of the following commonly used two Neural IR models:
\begin{enumerate}
    \item \textbf{Deep Relevance Matching Model} (DRMM): this model puts more emphasis on the relevance (both semantic and lexical) matching of the query rather than on exclusively semantic matching. It considers three crucial factors of the "handling of the exact matching signals, query term importance, and diverse matching requirements" \cite{guo2016deep}. 
    \item \textbf{Deep Semantic Similarity Model} (DSSM): based on the Siamese network architecture, DSSM has the main focus in comparing cosine similarities of the vector representations of a query and the document, where vector representations are learned using Deep fully-connected layers \cite{huang2013learning}. Originally such a model was only used for short text matching tasks (for example, matching questions with the most relevant answers); however, later, DSSM proved to be useful for tasks involving documents containing long texts, thus being a perfect choice for IR related tasks \cite{mitra2017learning}.
    
\end{enumerate}

\section{Experimental Results}

\subsection{Experimental Setup}
\partitle{Data.} We use the publicly available Qulac\footnote{\url{http://www.github.com/aliannejadi/qulac}} dataset, which is built based on the TREC Web Track 2009-2012, for our experiments. For keyword extraction experiments, we use the two datasets described in Section~\ref{sec:data}.

\partitle{Metrics.} We evaluate keyword extractions' performance in two ways, namely, the accuracy of extraction and end-to-end document retrieval. As for extraction accuracy, we use the following evaluation metrics: Precision, Recall, Average Tag Correct Identification (ATCI) \footnote{tests the quality of the overall assigned tags, by checking if the model has correctly assigned \textit{keyword} or \textit{not keyword} tag}, and Correct per Response Fill (CpRF) \footnote{It captures the ratio of fully correct and partially correct predictions to the total amount of sentences in the dataset (adopted from MUC-5)}. Also, we perform a human evaluation on the extracted keywords, where we ask the human annotators to score each extracted keyword from 1 to 5. Our IR evaluation follows the standard IR metrics, namely, Normalized Discounted Cumulative Gain at $k$ (nDCG@$k$), Precision at $k$ (P@$k$), Mean Reciprocal Rank (MRR), and Mean Average Precision (MAP).

\partitle{Statistical Significance.} We perform two-tailed paired t-test $p-value < 0.05$  to determine significant improvements on the IR metrics.

\subsection{Keyword Extractors}
To evaluate the performance of Keyword Extracting Neural Networks, two methods have been used. The first one relies on \textit{test dataset accuracy}, while the second one is a \textit{human evaluation} method.

\partitle{Test dataset accuracy.} Table \ref{tab:kwds_extractors_performance} shows the performance of the Keyword Extractors 
We also created a simple "Non-Neural approach" to serve as a baseline. This method operates in a very elementary way: the word frequencies were calculated from the training dataset. Using a brute force approach, the optimal frequency threshold was found, which maximizes the correct identification of tagged keywords (if the frequency of a certain word is below the threshold, the word is assigned to be a keyword). If the word has not been seen before, it is automatically assigned as the keyword as it is considered rare enough as it has not appeared in the training corpus.
In Table~\ref{tab:kwds_extractors_performance}, we see that BERT seems to be an ultimate solution to the keyword extraction problem, as the model achieves a much better test set performance than other keyword extractors.

\begin{table}[t]
    \centering
    \vspace{-3mm}
    \caption{Performance of keyword extractors on the test set}
    \begin{tabular}{lLLLLL}
       \toprule
        & Precision & Recall & ATCI & CpRF & Human Evaluation \\
        \midrule
        Non-Neural & 0.4572 & 0.7186 & 0.6216 & 0.4873 & -\\
        seq2seq & 0.6744 & 0.7795 & 0.8311 & 0.5863 & \textbf{4.268}\\
        BERT & \textbf{0.8636} & \textbf{0.8792} & \textbf{0.9178} & \textbf{0.7612} & 3.964\\
        \bottomrule
    \end{tabular}
    \vspace{-5mm}
    \label{tab:kwds_extractors_performance}
\end{table}

\partitle{Human Evaluation.} While the testing set evaluates the models’ performances in a similar training environment, it is also essential to test whether the extracted keywords would suit human judgments. To address this question, Google’s Query Wellformedness dataset has been used \cite{faruqui2018identifying}. 
Judges were asked to select the least possible number of keywords given a sentence and rate the relevance of the keywords chosen by the keyword extractor on a scale from 1 to 5, where 5 is the best score. For the latter part, we asked judges to imagine themselves in a situation where they have to answer the question based on only the keywords provided. In the scenario that the Neural Network’s selected keywords were doubtful, we asked the judges to use Google’s Search Engine and plug the keywords to see if sufficiently good results were obtained.

\begin{figure}[t]
    \centering
    \includegraphics[width=0.4\textwidth]{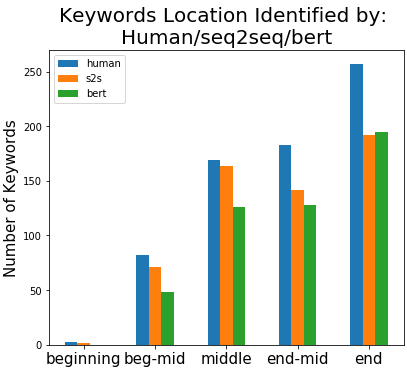}
    \vspace{-2mm}
    \caption{Performance of Keyword Extractors according to the human evaluation for various keywords positions in sentences.}
    \vspace{-5mm}
    \label{fig:human_eval}
\end{figure}

As can be observed from Table \ref{tab:kwds_extractors_performance}, Sequence-to-Sequence (Seq2seq) models appear to retrieve better keywords than BERT as the scores given to the model by the judges are higher.
Additionally, it is essential to focus on words that the judges selected, the Seq2seq and BERT models alike, to describe the reasoning of the classifiers and compare it to the experts’ judgment. We explore the locations in which the keywords tend to appear in the sentences more often. As clearly seen from Figure \ref{fig:human_eval}, keywords appear to be located closer to the end of the sentence. Neural Networks have correctly learned this trend. However, it can be noted that, in general, the models tend to underestimate the number of keywords in a sentence. 

Both evaluation approaches (test set and human evaluation) give interesting insights as we can clearly see that BERT learned better keywords that lead to the best document retrieval. In contrast, according to the Human evaluation dataset, Sequence-to-Sequence was able to retrieve more relevant keywords. Therefore, we also study the impact of both models on end-to-end IR performance to see how they eventually affect IR performance.

\subsection{Neural IR Models}

The performance of the Neural IR Models is presented in Table \ref{tab:IR_results_table}. As can be observed, in the case of the DRMM Neural IR model usage, the models provided with keywords have achieved a similar performance and have outperformed the Non-Neural IR model. Interestingly, the DRMM supplied with original conversational sequences was able to show the best performance concerning other DRMM models.

Looking at the DSSM, we can observe that providing keywords using BERT or the Sequence-to-sequence architecture yields much better results than when using original conversational sequences or the Non-Neural model (the last two achieved relatively similar performances). DSSM models that have used keywords have achieved the overall best retrieval performance.

\begin{table}[t]
    \centering
    \vspace{-3mm}
    \caption{Performance of the IR models. Bold values denote the best performance for each metric. $\dagger$ denotes significant improvements compared to the \textit{DRMM-orig} model using two-tailed paired t-test with $p < 0.05$.}
    \begin{adjustbox}{width=1\textwidth, rotate=0}
        \begin{tabular}{llrrrrrr}
        \toprule
              &            &    nDCG@3 &    nDCG@5 &  P@3 &       P@5 &       MRR &       MAP \\
        \midrule
        DRMM & orig &  0.3039 &  0.3304  &  0.2502 &  0.2289  &  0.3925 &  0.3433 \\
              & s2s &  0.2953 &  0.3117 &  0.2491 &  0.2200  &  0.3759 &  0.3272 \\
              & bert &  0.2939 &  0.3126 &  0.2510 &  0.2215 &  0.3761 &  0.32586 \\
              & non-neural &  0.2117 &  0.2328 &  0.1812 &  0.1718 &  0.3027 &  0.2529 \\
        \midrule
        DSSM & orig &  0.2117 &  0.2328 &  0.1812 &  0.1718 &  0.3027 &  0.2529 \\
              & s2s &  0.3194$^\dagger$ &  \textbf{0.3415}$^\dagger$ & \textbf{0.2710}$^\dagger$ &  \textbf{0.2383}$^\dagger$ &  0.3914$^\dagger$ &  0.3420 \\
              & bert &  \textbf{0.3231}$^\dagger$ &  0.3386$^\dagger$  &  0.2677$^\dagger$ &  0.2345$^\dagger$ & \textbf{0.4079}$^\dagger$ &  \textbf{0.3476} \\
              & non-neural &  0.2117 &  0.2328 & 0.1812 &  0.1718 &  0.3027 &  0.2529 \\

        \bottomrule
        \end{tabular}
        \vspace{-5mm}
    \end{adjustbox}
    
    \label{tab:IR_results_table}
\end{table}

\partitle{Keywords Extractor Influence on IR model.}
Another interesting insight is provided by considering how well the Neural IR model performs concerning the effectiveness of the Keyword Extractor. In this case, we are interested in the \textit{precision} of keywords provided by the Sequence-to-sequence Keyword Extractor and how the produced keywords impact the performance of the DSSM model.

\begin{table}[t]
    \centering
    \vspace{-3mm}
    \caption{Performance of Neural IR models on dataset depending on the precision of Sequence-to-sequence Keyword Extractor}
    \begin{tabular}{lLLL}
       \toprule
        & NDCG@5 & P@5 & MAP \\
        \midrule
        Low Precision & 0.2445 & 0.1557 &0.2476 \\
        Medium Precision & 0.4399 & 0.3127 & 0.4324 \\
        High Precision & 0.4503 & 0.3502 & 0.4595 \\
        \bottomrule
    \end{tabular}
    \vspace{-5mm}
    \label{tab:prec_kwdExtr_IRmodel}
\end{table}

As we see in Table~\ref{tab:prec_kwdExtr_IRmodel}, the premise is that \textit{the better the quality of the produced keywords, the better the IR model will perform}. It is also interesting to see how much the Neural IR model will benefit from high-quality keywords.  
First, we start with the test dataset ordered by the relevance scores assigned by the Neural IR models. The next step is to split the dataset into three sub-parts, based on the precision of the keywords obtained from the Keyword Extractor's query-question-answer sequences. 

\begin{equation}
P(k) = 
    \begin{cases}
    \text{low} & 0 \le P(k) < 0.5 \\
    \text{medium} & 0.5 \le P(k) < 0.75 \\
    \text{high} & \text{otherwise}
    \end{cases}
\end{equation}

Where $P(k)$ is the precision of the keyword extractor on the sentence $k$. 

Table \ref{tab:prec_kwdExtr_IRmodel} shows how much the quality of the keywords influences the document retrieval of DSSM models. As it can be observed, the model has managed to rank the documents much better when the supplied keywords had a high relevance.

\section{Conclusions}

This research studied the application of keywords in the context of conversational IR is going to be advantageous. For this purpose, we created two keyword extraction datasets and studied two types of Keyword Extractor, one based on a seq2seq architecture and the other based on BERT. We tested the keyword extraction performance based on keyword extraction, as well as end-to-end document retrieval performance. To do so, we test the performance on two state-of-the-art neural IR models, namely, DRMM and DSSM.
We showed that Neural IR models supplied with keywords from conversational communications with users improve the relevance of retrieved documents through experimental results. In addition, we showed that the higher the Keyword Extractor's precision, the better is the performance of the DSSM IR model. 

For further work, it would be interesting to train the Neural Networks on a newly created dataset manually labeled by humans. Likely, the keyword dataset creation approach, which we proposed in this paper, misses some important keywords that humans will identify easily. Moreover, we plan to use the same data and technique to improve conversational search in the context of unified mobile search~\cite{DBLP:conf/sigir/AliannejadiZCC18,DBLP:conf/cikm/AliannejadiZCC18,DBLP:journals/tois/AliannejadiZCC21} where users interact with their smartphones through a unified voice-based interface.

\section*{Acknowledgement}
The work constitutes part of the master thesis of Oleg Borisov at the Universit\'a della Svizzera italiana (USI) on conversational search.

\bibliography{bibliography.bib}

\end{document}